\documentclass[10pt,letterpaper,twocolumn]{article}

\usepackage{graphicx} 
\usepackage[utf8]{inputenc}
\usepackage[T1]{fontenc}
\usepackage{float}
\usepackage{hyperref}
\usepackage{colortbl}
\usepackage{longtable}
\usepackage{tabularx}
\usepackage{array}
\usepackage{booktabs}
\usepackage{enumitem}
\usepackage{fancyhdr}
\usepackage{lastpage}
\usepackage{titlesec}
\usepackage{tocloft}
\usepackage{amsmath}
\usepackage{amsfonts}
\usepackage{tikz}
\usepackage{pgfplots}
\usepackage{listings}
\usepackage{url}
\usepackage{forest} 
\usepackage{multirow}
\usepackage{placeins}
\usepackage{adjustbox}
\usepackage{comment}
\usepackage{xspace}
\usepackage[usenames,dvipsnames]{xcolor}

\usepackage{diagbox}

\usepackage{url}
\usepackage{hyperref}
\usepackage{xurl} 

\date{}

\newcommand{\name}{\texttt{MATRA}\xspace}

\title{\name: Modeling the Attack Surface of Agentic AI Systems - OpenClaw Case Study}

\author{
  Tim Van hamme, \quad
  Thomas Vissers, \quad
  Javier Carnerero-Cano, \\
  Mario Fritz, \quad
  Emil C. Lupu,\quad
  Lieven Desmet,  \quad
  Dinil Mon Divakaran\footnote{Tim Van hamme, Thomas Vissers and Lieven Desmet are with DistriNet, KU Leuven, 3001 Leuven, Belgium (\texttt{\{tim.vanhamme, thomas.vissers, lieven.desmet\}@kuleuven.be}), Javier Carnerero-Cano with IBM Research (\texttt{javier.cano@ibm.com}), Mario Fritz with CISPA Helmholtz Center for Information Security (\texttt{fritz@cispa.de}), Emil C. Lupu with Imperial College London (\texttt{e.c.lupu@imperial.ac.uk}) and Dinil Mon Divakaran with A*STAR Institute for Infocomm Research (\texttt{dinil\_divakaran@a-star.edu.sg}).}
  \thanks{Accepted for presentation at the 5th International Workshop on Designing and Measuring Security in Systems with AI (DeMeSSAI 2026), co-located with the 11th IEEE European Symposium on Security and Privacy (EuroS\&P 2026), Lisbon, Portugal, July 10, 2026.}
}

\begin{document}

\maketitle

\begin{abstract}
LLMs are increasingly deployed as autonomous agents with access to tools, databases, and external services, yet practitioners (across different sectors) lack systematic methods to assess how known threat classes translate into concrete risks within a specific agentic deployment. We present \name, a pragmatic threat modeling framework for agentic AI systems that adapts established risk assessment methodology to systematically assess how known LLM threats translate into deployment-specific risks. \name begins with an asset-based impact assessment and utilizes attack trees to determine the likelihood of these impacts occurring within the system architecture. We demonstrate \name on a personal AI agent deployment using OpenClaw, quantifying how architectural controls such as network sandboxing and least-privilege access reduce risk by limiting the blast radius of successful injections.
\end{abstract}

\section{Introduction}

Large Language Models (LLMs) are increasingly being embedded into enterprise software systems that ingest untrusted content and produce operational decisions, recommendations, or even executable plans. In these \emph{LLM-integrated} applications, security failures are increasingly driven not only by the behavior of the model in isolation, but by how model outputs are connected to, e.g., privileged tools, databases, retrieval pipelines, and downstream services \cite{meta2025ruleoftwo}.

Recently, researchers have demonstrated a growing catalogue of LLM safety and security failures, including jailbreaking and prompt injection, and the ecosystem has responded with guardrails \cite{protectai2024deberta, google2025modelarmor, chennabasappa2025llamafirewall}, automated red-teaming methods \cite{mehrotra2024tree, derczynski2024garakframeworksecurityprobing, zou2023universal}, and benchmarks for attacks and defenses~\cite{ debenedetti2024agentdojo, chao2024jailbreakbench}. Yet, a persistent disconnect remains between vulnerability-centric evaluations and the concrete risks faced by enterprises deploying LLMs in real business workflows, where harm is realized only after a model failure propagates through application logic and connected assets~\cite{htb2025echoleak, willison2025summerjohann}.

Prompt injection is now widely recognized as a practical and difficult-to-mitigate attack~\cite{nasr2025attackermovessecondstronger} in LLM-integrated systems. Beyond direct user-to-model injections, \emph{indirect} prompt injection can be delivered via retrieved documents or tool outputs, allowing adversaries to influence an application without a direct interface~\cite{10.1145/3605764.3623985}. This risk is amplified in \emph{agentic} workflows, where an LLM iteratively plans and executes tool calls: agentic pipelines blur the boundary between \emph{instructions} and \emph{data} \cite{zverev25iclr}, making agents susceptible to being steered by untrusted contents they process as part of the task. Recent benchmarks show that attacks can succeed at multiple stages of the agentic loop and existing defenses provide limited and inconsistent protection~\cite{nasr2025attackermovessecondstronger, prompt-tool-shi-NDSS2026, ObliInjection-NDSS-2026}.

A fundamental constraint underlies these threats: prompt injection is functionally equivalent to delivering new instructions to the LLM, much as code injection delivers executable instructions to a computer. Existing defenses can reduce the probability of successful injection or potentially detect the injection, but cannot eliminate it~\cite{nasr2025attackermovessecondstronger, pandya2025may,abdelnabi25satml}. Moreover, the attack surface extends beyond a single prompt-response interaction---persistent components such as system prompts, long-term memory, and retrieval-augmented generation (RAG) knowledge bases can be targeted for long-horizon influence and cross-session compromise~\cite{goodin2025geminiinjection}.

In response, the community has produced valuable taxonomies and governance guidance for GenAI risks, including the OWASP Top 10 for LLM Applications~\cite{owasp2025llmtop10}, 
the OWASP Agentic AI Threats and Mitigations guide~\cite{owasp2025agentic}, as well as NIST's profile for generative AI risks~\cite{nist2024gairmf}. However, these resources primarily describe \emph{what} classes of issues exist. Enterprises still lack a systematic framework to (i)~ground threats in a specific deployed workflow, (ii)~trace paths from impact scenarios to concrete attack vectors across tools, memory, and retrieval components, and (iii)~prioritize mitigations based on business impact.

We address this gap with \name\footnote{Mātrā in Sankskrit denotes measure, quantify, etc.}(\underline{M}odeling \underline{A}ttacks and \underline{T}hreats for \underline{R}isk \underline{A}ssessment), an impact-driven threat and risk modeling framework for agentic AI systems. \name is inspired by NIST SP 800-30 risk assessment concepts \cite{nist800-30} and models system risks by constructing attack trees that connect CIA-based impact scenarios to attacker objectives, techniques, and architecture-specific vectors. \name captures not only on adversarial manipulation but also non-adversarial pathways in which unexpected LLM behavior combines with autonomous tool execution to produce high-impact outcomes.

We demonstrate \name on OpenClaw, an open-source \emph{single-agent, multi-tool} personal assistant architecture that integrates messaging channels, high-privilege tools (shell, file I/O, web), and persistent memory \cite{steinberger2025openclaw}. Our case study shows how architectural patterns determine the feasible attack surface and how controls can be evaluated for risk reduction from two complementary perspectives: (i) the weakest-link (easiest-path) risk; and (ii) the full-attack surface exposure across feasible vectors.

\section{\name: Threat and Risk Modeling Framework}
\label{sec:matra-framework}

\begin{figure*}
    \centering
    \includegraphics[width=0.85\linewidth]{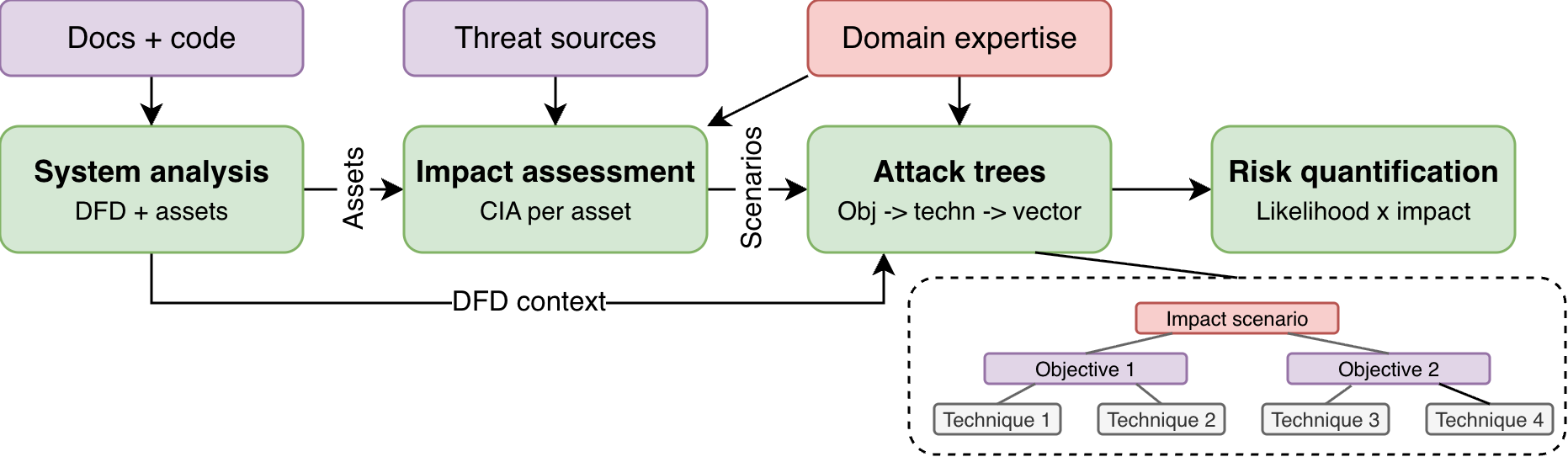}
    \caption{\name framework overview. System properties and threat sources are collected from the client. Assets identified from system documentation feed into a stakeholder-driven business impact assessment, which produces impact scenarios. A data flow diagram (DFD), combined with known attack techniques from established catalogs, informs the construction of attack trees that decompose each impact scenario into objectives, techniques, and architecture-specific vectors.}
    \label{fig:matra-framework}
\end{figure*}

\name framework is inspired by NIST Special Publication 800-30, ``Guide for Conducting Risk Assessments’’~\cite{nist800-30}. We demonstrate it using a \textbf{semi-quantitative approach} to risk assessment, where structured scales translate expert judgment into ratings that enable relative risk comparison and prioritization while remaining accessible to decision-makers. Other assessment approaches, e.g., fully quantitative or purely qualitative, can be used within \name.

The assessment is \textbf{impact-oriented}: it begins with the identification of critical assets and the business impacts of concern, then traces backward to the threat events and sources that could cause those impacts. This ensures that vulnerabilities and adversarial actions are examined in the context of their potential to create unacceptable outcomes rather than in isolation.
Figure~\ref{fig:matra-framework} summarizes the framework’s information flow. A practitioner proceeds through four stages: system analysis, impact assessment, attack tree construction, and risk quantification.

\subsection{System Analysis}
\label{subsec:system-analysis}

The assessment begins by collecting system properties from the stakeholder: architecture documentation, source code, deployment configurations, and any existing security documentation. From these inputs, the practitioner constructs a \textbf{data flow diagram (DFD)} that maps processes, data stores, external actors, and trust boundaries.

The DFD serves two purposes. First, it identifies the system’s \textbf{assets}, i.e., the critical resources requiring protection, such as databases, credentials, user data, and integration interfaces. Systematically identifying assets is a prerequisite for the impact assessment that follows, as it enables the direct mapping of potential impact scenarios to specific system components. Second, the DFD exposes the system’s communication paths and trust boundaries, which later inform the identification of concrete attack vectors.

\subsection{Impact Assessment}
\label{subsec:impact-assessment}

With assets identified, the practitioner selects the \textbf{threat sources} relevant to the deployment. Threat sources are entities or conditions capable of exploiting vulnerabilities within the system; ranging from nation-state actors and competitors to insiders and accidental causes (see Appendix~\ref{sec:appendix-threat-actors} for the full taxonomy adapted from NIST SP~800-30~\cite{nist800-30}). Only threat sources with credible motivation and a viable access path to the system under review are retained for analysis; this scoping step filters out actors that lack intent or a realistic targeting path.

For each retained threat source, or per threat source category when impact is independent of the attacker's identity, the stakeholder assesses \textbf{business impact} across the CIA triad---confidentiality, integrity, and availability---for every asset in scope:

\emph{``What is the business impact if this asset’s [confidentiality / integrity / availability] is compromised by [threat source]?’’}

\noindent Impact is rated on a three-grade scale: \textbf{Low}: Disruption, financial loss, or reputation harm is limited. Mitigation is quick and recovery is straightforward. \textbf{Moderate}: Disruption or losses are noticeable. Recovery requires effort and additional resources. \textbf{High}: Major disruption of operations, significant financial losses or reputation harm. Recovery is prolonged or potentially impossible.

From this matrix, the practitioner derives \textbf{impact scenarios} that are distinct, architecturally viable adverse outcomes, rated at the highest assessed level across threat sources. Each impact scenario names a specific asset, a CIA dimension, and the concrete harm (e.g., ``customer PII exfiltrated from the production database’’).

\subsection{Attack Tree Construction}
\label{subsec:attack-tree-construction}

To connect impact scenarios to concrete threats, the practitioner builds \textbf{attack trees}~\cite{schneier-attack-trees}---hierarchical diagrams that decompose high-level impacts into controllable, architecture-specific attack paths. The nodes of an attack tree are:

\begin{itemize}
  \item \textbf{Root:} The \emph{impact scenario}. Focusing on the most consequential impacts helps manage scale.
  \item \textbf{Attack Objectives:} One or more high-level goals that must be achieved to realise the impact (e.g., LLM behavioural manipulation combined with data exfiltration).
  \item \textbf{Attack Techniques:} Methods to achieve an objective such as indirect prompt injection.
  \item \textbf{Attack Vectors:} Concrete instances of a technique in the system architecture, e.g., indirect prompt injection via a support email.
\end{itemize}

The practitioner constructs trees in three stages. First, each impact scenario is decomposed into one or more attack objectives. Second, well-established LLM-specific threat repositories from, say, OWASP~\cite{owasp2025llmtop10,  owasp2025agentic, owasp2026agentic}, MITRE~\cite{mitre2025atlas}, NIST~\cite{nist2024gairmf}~ CSA \cite{csa2025aicm}, and Cisco~\cite{cisco2025aiframework} are consulted alongside the DFD and system model to select the most relevant attack techniques and to identify existing security controls. Third, each technique is mapped to the concrete attack vectors present in the application architecture, informed by the different paths in DFD and the trust boundaries.

This method ensures that each attack tree reflects both recognised industry threats and the specific architecture of the system under review.

\subsection{Risk Quantification}
\label{subsec:risk-quantification}
 
For each attack vector, likelihood is derived in two steps.
First, \emph{capability fit} compares the threat source's technical capability with the skill level required for the vector, using Table~\ref{table:capability-fit} (Appendix~\ref{sec:appendix-risk-matrices}); because the scoping step in Section~\ref{subsec:impact-assessment} already filters for intent and targeting, capability remains the discriminating factor at the vector level.
For non-adversarial threats, capability does not apply; instead, each vector receives an \emph{inherent manifestation likelihood} reflecting the probability of the event occurring independently of any actor (e.g., hallucination probability).
Second, \emph{residual success likelihood} estimates the probability that the vector succeeds given existing security controls; a vector with no applicable controls receives a high residual likelihood, while effective controls reduce it.
The overall vector likelihood is obtained by combining these two values via the combination matrix (Table~\ref{table:combined-score}, Appendix~\ref{sec:appendix-risk-matrices}).
 
Within each objective the most favourable (highest) vector likelihood is chosen, reflecting that an attacker will pursue the path of least resistance. Across objectives the overall scenario likelihood is the weakest link, i.e., the minimum of those maxima, because every objective must succeed for the impact to occur.
 
An impact scenario's risk is $\text{Risk} = \text{Likelihood} \times \text{Impact}$, where qualitative levels (low, moderate, high) combine into a final rating from very low to very high via Table~\ref{table:risk-calculation-matrix} (Appendix~\ref{sec:appendix-risk-matrices}).
We draw mitigation decisions from two complementary perspectives: (i)~\emph{weakest-link risk}, which measures the single easiest attack path an adversary is most likely to exploit first (or that is most likely to manifest non-adversarially); and (ii)~\emph{full attack surface}, which captures total exposure across all feasible vectors, revealing cases where multiple comparable paths create material risk even if no single path dominates.
Using both perspectives ensures that defences address not only the most obvious entry point but also the full spectrum of exploitable opportunities.

\section{Use Case: Personal AI Agent}

We demonstrate \name on OpenClaw, an open-source personal AI agent that follows a \emph{single-agent, multi-tool} architecture: one LLM agent with simultaneous access to shell execution, file I/O, web browsing, cross-channel messaging, and persistent memory~\cite{steinberger2025openclaw}. We model a solo developer using OpenClaw as a productivity tool for their SaaS booking platform, with an email plugin for customer support. The full deployment scenario, data flow diagrams, asset inventory, threat source selection, and complete attack trees are provided in Appendix~\ref{sec:appendix-openclaw-supporting}.

From the full assessment, we select one threat source--impact combination to illustrate \name's end-to-end process: a \emph{malicious customer} (threat source) targeting the \emph{confidentiality of customer PII} (business impact). This combination is compelling because the threat source has a natural, legitimate access channel, i.e., customer support email, and requires no technical sophistication beyond crafting a convincing message with embedded instructions.

\subsection{Deep Dive: Malicious Customer $\times$ Customer PII Confidentiality}
\label{sec:openclaw-deepdive-exfil}

In this scenario (IS6 in Table~\ref{table:openclaw-impact-scenarios}), a malicious customer exploits the agent's email channel to exfiltrate other customers' PII (personally identifiable information) and booking records, following the indirect prompt injection pattern demonstrated by Zhao et al.~\cite{zhao2025openclaw}. Two objectives must jointly succeed: \emph{LLM behavioral manipulation} (redirecting the agent's actions) and \emph{data exfiltration} (transmitting customer records outside the trust boundary).

The attacker sends a support email with hidden instructions embedded in an attachment (e.g., via HTML comments or zero-width Unicode characters). When the agent processes the email, the injected instructions direct it to query customer records via the \texttt{exec} tool and exfiltrate the results through tool-based channels (\texttt{curl}, email) or rendered-markup channels (embedding data in a markdown image URL that the Telegram client auto-fetches). The entire chain executes within a single agent turn.

\paragraph{Attack Tree with Likelihood Scores}

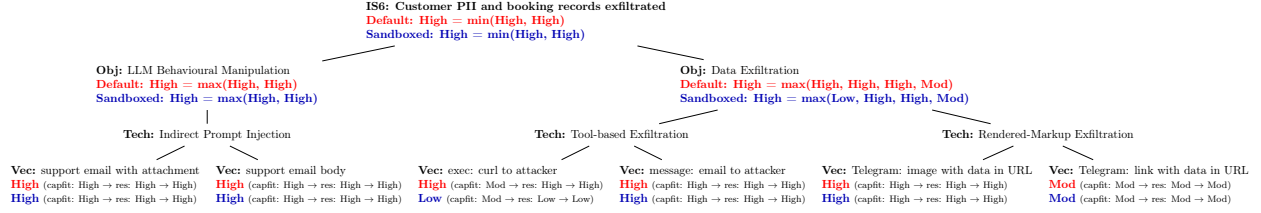
\begin{figure*}[]
\centering
\resizebox{\linewidth}{!}{%
\begin{forest} for tree={align=left}
[\textbf{IS6: Customer PII and booking records exfiltrated}\\
  \textcolor{red}{\textbf{Default: High = min(High, High)}}\\
  \textcolor{blue!70!black}{\textbf{Sandboxed: High = min(High, High)}}
  [\textbf{Obj:} LLM Behavioural Manipulation\\
    \textcolor{red}{\textbf{Default: High = max(High, High)}}\\
    \textcolor{blue!70!black}{\textbf{Sandboxed: High = max(High, High)}}
    [\textbf{Tech:} Indirect Prompt Injection
      [\textbf{Vec:} support email with attachment\\
        \textcolor{red}{\textbf{High}} \footnotesize{(capfit: High $\to$ res: High $\to$ High)}\\
        \textcolor{blue!70!black}{\textbf{High}} \footnotesize{(capfit: High $\to$ res: High $\to$ High)}
      ]
      [\textbf{Vec:} support email body\\
        \textcolor{red}{\textbf{High}} \footnotesize{(capfit: High $\to$ res: High $\to$ High)}\\
        \textcolor{blue!70!black}{\textbf{High}} \footnotesize{(capfit: High $\to$ res: High $\to$ High)}
      ]
    ]
  ]
  [\textbf{Obj:} Data Exfiltration\\
    \textcolor{red}{\textbf{Default: High = max(High, High, High, Mod)}}\\
    \textcolor{blue!70!black}{\textbf{Sandboxed: High = max(Low, High, High, Mod)}}
    [\textbf{Tech:} Tool-based Exfiltration
      [\textbf{Vec:} exec: curl to attacker\\
        \textcolor{red}{\textbf{High}} \footnotesize{(capfit: Mod $\to$ res: High $\to$ High)}\\
        \textcolor{blue!70!black}{\textbf{Low}} \footnotesize{(capfit: Mod $\to$ res: Low $\to$ Low)}
      ]
      [\textbf{Vec:} message: email to attacker\\
        \textcolor{red}{\textbf{High}} \footnotesize{(capfit: High $\to$ res: High $\to$ High)}\\
        \textcolor{blue!70!black}{\textbf{High}} \footnotesize{(capfit: High $\to$ res: High $\to$ High)}
      ]
    ]
    [\textbf{Tech:} Rendered-Markup Exfiltration
      [\textbf{Vec:} Telegram: image with data in URL\\
        \textcolor{red}{\textbf{High}} \footnotesize{(capfit: High $\to$ res: High $\to$ High)}\\
        \textcolor{blue!70!black}{\textbf{High}} \footnotesize{(capfit: High $\to$ res: High $\to$ High)}
      ]
      [\textbf{Vec:} Telegram: link with data in URL\\
        \textcolor{red}{\textbf{Mod}} \footnotesize{(capfit: Mod $\to$ res: Mod $\to$ Mod)}\\
        \textcolor{blue!70!black}{\textbf{Mod}} \footnotesize{(capfit: Mod $\to$ res: Mod $\to$ Mod)}
      ]
    ]
  ]
]
\end{forest}%
}
\caption{Attack tree for IS6 (customer data exfiltration; refer Table~\ref{table:openclaw-impact-scenarios}),
  assessed for threat source: \emph{malicious customer} (capability: Low).
  Likelihood scores shown under default configuration (red) and with Docker sandboxing (blue).
  Vector labels follow the format \emph{capfit: capability fit $\to$ res: residual success likelihood $\to$ combined likelihood}
  (see Tables~\ref{table:capability-fit} and~\ref{table:combined-score}).
  Sandboxing blocks \texttt{exec}-based network access (res: Low for curl) but does not affect
  message-based or rendered-markup vectors, which operate outside the container (res: High);
  the link vector's requirement for a user click partially reduces success likelihood (res: Mod).}
\label{fig:openclaw-at-is6}
\end{figure*}

We assess this scenario 
under two control configurations:
(a)~default unsandboxed deployment, and
(b)~Docker sandboxing enabled (\texttt{network=none}).
The resulting attack tree is shown in Figure~\ref{fig:openclaw-at-is6}.

\paragraph{Risk Assessment}

Impact is rated High due to the regulatory implications of customer PII disclosure. Without sandboxing, the scenario reaches Very~High risk~(9). Docker sandboxing with \texttt{network=none} blocks \texttt{exec}-based exfiltration (\texttt{curl}) but does not reduce the overall exfiltration objective: the message tool runs on the gateway host outside the container, so email-based exfiltration remains viable, and the agent's Telegram response can embed queried data in a markdown image URL that the client auto-fetches (zero-click). The scenario therefore remains at Very~High~(9) even with sandboxing.

Reducing IS6 to Moderate~(3) requires controls that address the surviving exfiltration vectors: an email allow-list that restricts outbound recipients, and output sanitisation that strips or rewrites URLs containing encoded data from agent responses before they reach the messaging channel. With both controls in place, all exfiltration vectors are reduced to Low, yielding Moderate risk. The injection objective remains at High regardless---customer emails are \emph{expected} input that the agent is designed to process.

\subsection{Key Observations}

\paragraph{Prompt Injection and Architectural Defense}

Prompt injection is functionally equivalent to delivering new instructions to the LLM: like code injection in traditional systems, the injected payload executes with whatever capabilities the host process provides. Because current defenses can reduce but not eliminate the probability of successful injection \cite{nasr2025attackermovessecondstronger}, the impact of an attack is ultimately determined by what the agent can \emph{do} with the tools, integrations, and permissions available to it.

This has a direct consequence for risk mitigation. Rather than relying primarily on input-level defenses (prompt injection detectors, guardrails), the most effective controls are \emph{architectural}: sandboxing that limits network access, allow-listing that restricts which tools and commands the agent may invoke, and least-privilege configurations such as read-only database roles. The IS8 analysis (Appendix~\ref{sec:appendix-deepdive-db}) demonstrates this clearly: a read-only database role eliminates the action execution objective entirely, reducing risk from High~(6) to Moderate~(3). However, the IS6 analysis reveals that no single architectural control is sufficient when multiple exfiltration channels exist---container sandboxing blocks \texttt{exec}-based vectors but leaves message-based and rendered-markup vectors unaffected. Effective defense requires \emph{layered} architectural controls that address each exfiltration category.

\paragraph{Non-Adversarial Threats}

Autonomous tool execution creates a qualitatively new risk category: high-impact scenarios requiring no attacker. As demonstrated in the IS8 analysis (Appendix~\ref{sec:appendix-deepdive-db}), a hallucinated \texttt{UPDATE} with a missing predicate can corrupt customer booking records, trigger erroneous refunds, and produce incorrect reports---all from a routine developer request. In traditional systems, integrity impacts of this magnitude typically require adversarial action. In agentic systems, the LLM's imperfect reasoning combined with unrestricted tool access means that normal usage can trigger destructive operations. This has implications for risk frameworks: \emph{accidental} threat sources must be assessed with the same rigor as adversarial ones when the system has autonomous execution capabilities.

\section{Conclusions}

We presented \name, an impact-driven threat and risk modeling framework for agentic AI systems. \name provides a structured process comprising system analysis, stakeholder-driven impact assessment, attack tree construction informed by established threat catalogs, and risk quantification. They connect high-level business impacts to architecture-specific attack vectors. Inspired by NIST SP 800-30 and grounded in data flow analysis, \name enables practitioners to move beyond generic vulnerability lists toward deployment-specific risk prioritization.

Our case study on OpenClaw reinforces a central insight: because prompt injection cannot be eliminated \cite{nasr2025attackermovessecondstronger}, the most effective risk reduction comes from limiting what injected instructions can achieve through layered architectural controls via dedicated planner and executer LLMs \cite{debenedetti2025defeatingpromptinjectionsdesign, costa2025securingaiagentsinformationflow}, sandboxing, allow-listing, output sanitisation, and least-privilege configurations, each addressing different vector classes.

Several limitations point to future work. Attack tree completeness is bounded by an analyst's knowledge and the threat catalogs consulted. However, the hierarchical structure makes gaps localizable; when a new technique emerges, the affected objectives and impact scenarios are immediately identifiable. \name currently relies on expert judgment for likelihood assignment; we bound this subjectivity through deliberately coarse scales and auditable per-node ratings, but integrating empirical data from red-teaming benchmarks or operational logs could improve calibration. Furthermore, the min-max composition heuristic may underestimate risk when multiple moderate paths interact; the full attack surface perspective partially addresses this, but formal composition under uncertainty is a future direction. \name has been demonstrated on a single-agent architecture; applying it to multi-agent orchestrations with inter-agent communication would test its scalability and may require extensions to model cross-agent propagation paths. Finally, our assessment does not address privacy-specific threat modeling (e.g., LINDDUN\cite{SION2025104419}), which is an important complementary dimension for systems processing personal data.

 \paragraph{Acknowledgments} This research is partially funded by the Internal Funds KU Leuven, and by the Cybersecurity Research Program Flanders.

 This research is also supported by the National Research Foundation, Singapore, and the Cyber Security Agency of Singapore under the National Cybersecurity R\&D Programme (NCR25-CRPO-0001) and the CyberSG R\&D Programme Office (Award CRPO-GC5-ASTAR-001). Any opinions, findings, conclusions, or recommendations expressed in these materials are those of the author(s) and do not reflect the views of the National Research Foundation, Singapore, the Cyber Security Agency of Singapore, or the CyberSG R\&D Programme Office.

 \noindent Last but not the least, the authors thank Schloss Dagstuhl -- Leibniz Center for Informatics for hosting Seminar 25461 on ``Security and Privacy of Large Language Models,''\footnote{https://www.dagstuhl.de/25461} and its organizers and participants for stimulating discussions that sharpened this work.


\bibliographystyle{ieeetr}
\bibliography{bibliography}

\appendix

\section{Risk Quantification Matrices}
\label{sec:appendix-risk-matrices}

\begin{table}[h]
\centering
\small
\caption{Capability fit lookup: adversary capability (rows) versus vector skill requirement (columns).}
\begin{tabular}{c| c c c}
\toprule
 \diagbox[width=2.85cm]{\textbf{Capability}}{\textbf{Skill Req.}} & \textbf{Low} & \textbf{Mod} & \textbf{High} \\
\midrule
High  & High & High & High \\
Mod   & High & High & Mod \\
Low   & High & Mod & Low \\
\bottomrule
\end{tabular}
\label{table:capability-fit}
\end{table}

\begin{table}[h]
\centering
\small
\caption{Combination matrix for deriving overall vector likelihood from capability fit (rows) and residual success likelihood (columns).}
\begin{tabular}{c | c c c}
\toprule
 \diagbox[width=4.42cm]{\textbf{Capability}}{\textbf{Res. likelihood}} & \textbf{Low} & \textbf{Mod} & \textbf{High} \\
\midrule
Low  & Low & Low & Mod \\
Mod  & Low & Mod & High \\
High & Mod & High & High \\
\bottomrule
\end{tabular}
\label{table:combined-score}
\end{table}

\begin{table*}[]
\centering
\caption{Risk Assessment Matrix}
\setlength{\extrarowheight}{2mm}
\begin{tabular}{|c|c|>{\centering\arraybackslash}p{3cm}|>{\centering\arraybackslash}p{3cm}|>{\centering\arraybackslash}p{3cm}|}
\hline
& & \multicolumn{3}{c|}{\textbf{Impact}} \\
\cline{3-5}
& & \textbf{Low} & \textbf{Moderate} & \textbf{High} \\
\hline
\multirow{3}{*}{\rotatebox{90}{\textbf{Likelihood}}}
& \textbf{High} & \cellcolor{yellow!50} Moderate (3) & \cellcolor{orange!50} High (6) & \cellcolor{red!50} Very High (9) \\
\cline{2-5}
& \textbf{Moderate} & \cellcolor{green!30} Low (2) & \cellcolor{yellow!50} Moderate (4) & \cellcolor{orange!50} High (6)\\
\cline{2-5}
& \textbf{Low} & \cellcolor{blue!20} Very Low (1) & \cellcolor{green!30} Low (2) & \cellcolor{yellow!50} Moderate (3) \\
\hline
\end{tabular}
\label{table:risk-calculation-matrix}
\end{table*}

\section{Threat Source Taxonomy}
\label{sec:appendix-threat-actors}

\name adopts the threat source taxonomy from NIST SP~800-30~\cite{nist800-30}, which distinguishes \emph{adversarial} and \emph{non-adversarial} sources. Adversarial sources are characterised by capability, intent, and targeting; non-adversarial sources are characterised by the range of effects and inherent likelihood of occurrence. Table~\ref{table:threat-source-taxonomy} summarises the taxonomy with examples relevant to agentic AI deployments.

\begin{table}[h]
\centering
\small
\caption{Threat source taxonomy adapted from NIST SP~800-30~\cite{nist800-30}.}
\begin{tabularx}{\columnwidth}{l X}
\toprule
\textbf{Type} & \textbf{Examples} \\
\midrule
\multicolumn{2}{l}{\textbf{Adversarial}} \\
\quad Individual  & Outsider, insider, privileged insider \\
\quad Group       & Ad hoc, established \\
\quad Organisation & Competitor, supplier, partner, customer \\
\quad Nation-State & State-sponsored actor \\
\midrule
\multicolumn{2}{l}{\textbf{Non-adversarial}} \\
\quad Accidental  & User error, automated process failure \\
\bottomrule
\end{tabularx}
\label{table:threat-source-taxonomy}
\end{table}

The practitioner selects threat sources from this taxonomy based on credible motivation and a viable access path to the system under review (see Section~\ref{subsec:impact-assessment}). For example, an agentic system exposed to customer email may warrant a \emph{customer} (organisation-level adversarial) source, while a system with autonomous tool execution should always include an \emph{accidental} (non-adversarial) source to account for hallucination-driven impacts.

\section{OpenClaw: Full Case Study}
\label{sec:appendix-openclaw-supporting}

\subsection{Deployment Scenario}
\label{sec:appendix-deployment-scenario}

We model a realistic deployment in which a solo developer runs a small B2B SaaS product---an online booking platform---and uses OpenClaw on their personal workstation as a productivity tool. The product's PostgreSQL database serves the web application, a background worker for payment processing, and a reporting dashboard; it is the single source of truth for customer data, bookings, and invoices. The developer interacts with the agent primarily via Telegram, issuing day-to-day instructions such as ``check the latest sign-ups,'' ``why is customer X getting errors,'' or ``deploy the fix to staging.'' An email plugin allows the agent to read and reply to customer support messages on the developer's behalf. The workspace contains the product's source code, deployment scripts, and a \texttt{.env} file with production database credentials and third-party API keys.

This deployment is configured as follows:

\begin{itemize}
  \item \textbf{Messaging channels:} Telegram (primary interface for instructing the agent) and email (via community extension plugin, for reading and sending customer support email).
  \item \textbf{LLM provider:} Anthropic Claude, accessed via API key stored in \texttt{\textasciitilde/.openclaw/credentials/}.
  \item \textbf{Tools enabled:} \texttt{exec} (shell commands), \texttt{browser} (Playwright-based web automation), \texttt{read}/\texttt{write}/\texttt{edit} (file I/O), \texttt{message} (cross-channel messaging), \texttt{memory\_search}/\texttt{memory\_get} (long-term memory).
  \item \textbf{Database access:} The production PostgreSQL database, shared with the web application and background workers; the agent queries it by running \texttt{psql} commands via the \texttt{exec} tool.
  \item \textbf{Workspace:} \texttt{\textasciitilde/.openclaw/workspace/} containing product source code, configuration files, and a \texttt{.env} file with database credentials and API keys.
  \item \textbf{Sandboxing:} Off (default configuration).
\end{itemize}

This scenario is directly grounded in OpenClaw's documented capabilities and default configuration. The addition of email is motivated by its prevalence as a community extension and its role as a high-value attack surface demonstrated in recent security research~\cite{zhao2025openclaw}.

\subsection{Assets and Impact Scenarios}
\label{sec:appendix-assets}

Following the process in Section~\ref{subsec:system-analysis}, we identify seven assets from the DFD (Figure~\ref{fig:openclaw-dfd-overview}): API credentials~(A1), messaging accounts~(A2), workspace files~(A3), conversation history and memory~(A4), the agentic application~(A5), the production PostgreSQL database~(A6), and the platform's end users~(A7). The database is particularly critical: it is shared with the web application, payment processing worker, and reporting dashboard, making it the single source of truth for customer PII, bookings, and invoices. End users are an asset because the agent can send messages on the developer's behalf---an attacker who compromises the agent can use this trusted channel to phish customers for their credentials.

Applying the impact assessment process from Section~\ref{subsec:impact-assessment}, we assess each asset against the CIA triad for each threat source. The full impact matrix is provided in Table~\ref{table:openclaw-impact-matrix}. From this matrix, we derive nine distinct impact scenarios, presented in Table~\ref{table:openclaw-impact-scenarios}. Eight of the nine scenarios are rated High impact, reflecting the deployment's broad attack surface and the sensitivity of the connected assets.

\subsection{Threat Source Selection}
\label{sec:appendix-threat-source-selection}

Following the process in Section~\ref{subsec:impact-assessment}, we select four threat sources from the NIST taxonomy (Appendix~\ref{sec:appendix-threat-actors}, Table~\ref{table:threat-source-taxonomy}) whose capability and access realistically intersect with the deployment's attack surface:

\begin{itemize}
  \item \textbf{Malicious customer} (Organisation $\rightarrow$ Customer; low capability): exploits the support email channel---a natural, legitimate access path---to inject instructions into the agent's context. This is the most realistic adversarial threat source due to the combination of low barrier, direct access, and clear motivation.
  \item \textbf{Competitor} (Organisation $\rightarrow$ Competitor; moderate capability): targets the customer database or source code via email or poisoned web pages.
  \item \textbf{Opportunistic outsider} (Individual $\rightarrow$ Outsider; low--moderate capability): relies on indirect prompt injection via web content the agent encounters during browsing.
  \item \textbf{Accidental} (Non-adversarial; no capability assessment): particularly significant because the agent autonomously executes tool calls without human confirmation. LLM hallucinations can trigger high-impact outcomes---such as corrupted database records---without any malicious input.
\end{itemize}

\subsection{Attack Tree Construction}
\label{sec:appendix-attack-tree-overview}

Following the process in Section~\ref{subsec:attack-tree-construction}, we construct attack trees in three stages: impacts to objectives, objectives to techniques, and techniques to architecture-specific vectors. Each impact scenario decomposes into two objectives that must both succeed. For adversarial confidentiality impacts, these are \emph{LLM behavioural manipulation} (redirecting the agent's actions) and \emph{data exfiltration} (transmitting data outside the trust boundary). For integrity impacts, the second objective becomes \emph{unauthorized action execution}---reflecting OpenClaw's broad write access to shell, files, messaging, and database. For availability, it becomes \emph{excessive resource consumption}. Non-adversarial paths mirror this structure, with \emph{unexpected LLM behaviour} (hallucination) replacing deliberate manipulation.

Two architectural features shape the resulting trees. First, OpenClaw supports two categories of exfiltration techniques: \emph{tool-based exfiltration}, where the agent explicitly invokes a tool to transmit data (e.g., \texttt{exec: curl}, \texttt{message: email}, Figure~\ref{fig:openclaw-exfil-vectors}), and \emph{rendered-markup exfiltration}, where the agent embeds queried data in a URL within its Telegram response and the client auto-fetches the resource (Figure~\ref{fig:openclaw-rendered-exfil-vectors}). A critical distinction for sandbox effectiveness is that only \texttt{exec}-based tools run inside the Docker container; the message tool and browser run on the gateway host, meaning their exfiltration vectors survive container-level network isolation. Second, because Telegram and email are accessible to external parties, \emph{direct} prompt injection (Figure~\ref{fig:openclaw-dpi-vectors}) is a primary concern---unlike systems where the only input channel is an authenticated internal chat. Indirect prompt injection vectors are particularly numerous due to the agent's broad input surface (Figure~\ref{fig:openclaw-ipi-vectors}).

Not all vectors apply to all scenarios; only architecturally viable combinations are included in each tree. For example, exfiltration vectors do not appear in integrity trees, and expensive query vectors apply exclusively to the database availability scenario. This scoping ensures each tree is realistic and prevents double-counting in aggregate risk assessment. The full trees are provided in Appendix~\ref{sec:appendix-attack-trees}.

\subsection{Data Flow Diagrams}
\label{sec:appendix-dfds}

\begin{figure*}[t]
\centering
\resizebox{\linewidth}{!}{%
\begin{tikzpicture}[
  external/.style={draw, thick, rectangle, minimum width=28mm, minimum height=12mm,
                   align=center, font=\footnotesize},
  process/.style={draw, thick, circle, minimum size=22mm, align=center,
                  font=\footnotesize},
  datastore/.style={minimum width=30mm, minimum height=8mm,
                    align=center, font=\footnotesize,
                    inner sep=3pt,
                    append after command={
                      (\tikzlastnode.north west) edge[thick] (\tikzlastnode.north east)
                      (\tikzlastnode.south west) edge[thick] (\tikzlastnode.south east)
                    }},
  flow/.style={->, >=stealth, thick},
  biflow/.style={<->, >=stealth, thick},
  trustboundary/.style={draw, dashed, thick, rounded corners=8pt},
]


\node[process, minimum size=28mm] (gateway) at (0.5, 2) {Gateway\\+ Agent\\(LLM)};
\node[process, minimum size=22mm] (tools) at (0.5, -2.5) {Tool\\Engine};

\node[datastore] (config) at (0.5, 5) {Config \& Credentials};

\node[datastore] (workspace) at (-3.2, -0.5) {Workspace\\Files};
\node[datastore] (memory) at (-2.5, -5.2) {Memory \&\\Sessions};
\node[datastore] (postgres) at (3.5, -5.2) {PostgreSQL\\Database};

\node[external] (user) at (-7.5, 5) {User};
\node[external] (telegram) at (-7.5, 2) {Telegram};
\node[external] (email) at (-7.5, -2) {Email\\Server};

\node[external] (llmprovider) at (8.5, 1.5) {LLM Provider\\(Anthropic)};
\node[external] (web) at (8.5, -2.5) {Web\\(untrusted)};

\node[trustboundary,
      fit=(config)(memory)(postgres)(workspace)(gateway)(tools),
      inner sep=12mm,
      label={[anchor=north west, font=\footnotesize\bfseries]north west:
      Trust Boundary: Host Device}] (host) {};

\node[trustboundary, minimum width=44mm, minimum height=28mm,
      label={[anchor=south west, font=\scriptsize\bfseries]south west:
      Docker Sandbox}] (sandbox) at (0.5, -2.5) {};


\draw[flow] (user.east) -- ++(0.8,0) |- node[above, font=\scriptsize, pos=0.25]
      {via Telegram / email} (gateway.150);

\draw[biflow] (telegram.east) -- node[above, font=\scriptsize] {messages} (gateway.west);
\draw[biflow] (email.east) -| node[below, font=\scriptsize, pos=0.25] {email}
      (gateway.south west);

\draw[biflow] (gateway.east) -- node[above, font=\scriptsize] {prompt / response}
      (llmprovider.west);

\draw[flow] (config) -- node[right, font=\scriptsize] {load} (gateway);

\draw[biflow] (gateway) -- node[left, font=\scriptsize, pos=0.45] {tool calls} (tools);

\draw[biflow] (tools.170) -- node[above, font=\scriptsize, sloped, pos=0.4]
      {read/write} (workspace.south);

\draw[biflow] (tools.south west) -- node[below, font=\scriptsize, pos=0.4]
      {search/store} (memory.north east);

\draw[biflow] (tools.south east) -- node[below, font=\scriptsize, pos=0.4]
      {psql} (postgres.north west);

\draw[biflow] (tools.east) -- node[above, font=\scriptsize, sloped, pos=0.5]
      {HTTP} (web.west);

\end{tikzpicture}%
}
\caption{High-level DFD of the OpenClaw deployment scenario.
  The gateway process mediates all interactions between messaging channels,
  the LLM provider, and the tool engine. Bidirectional arrows on Telegram
  and email reflect that the gateway both receives inbound messages and sends
  outbound messages (the message tool runs on the gateway, outside the
  Docker sandbox). Only the tool engine (exec, file I/O) runs inside the
  optional sandbox container.}
\label{fig:openclaw-dfd-overview}
\end{figure*}
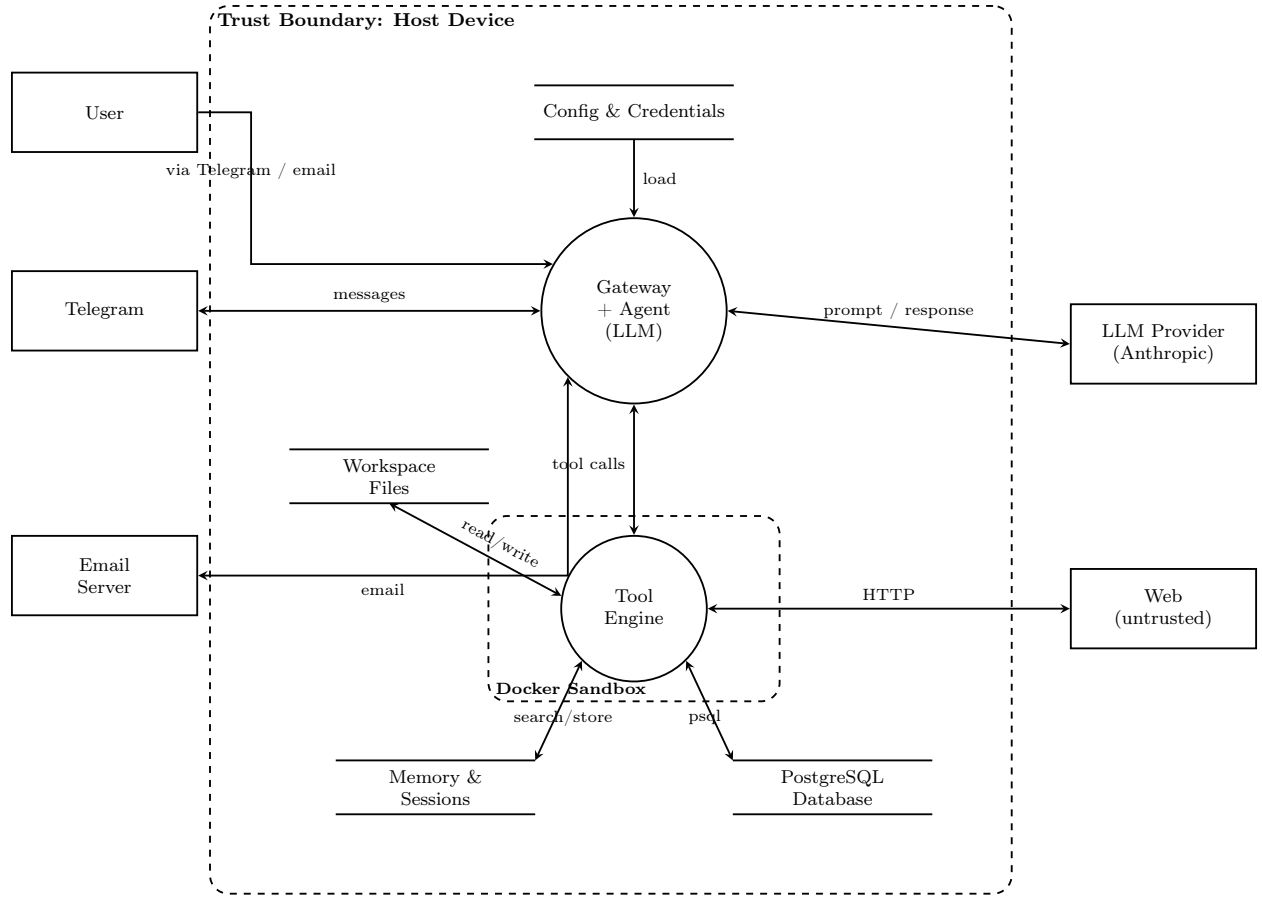

Figure~\ref{fig:openclaw-dfd-overview} presents the high-level data flow diagram for the OpenClaw deployment, showing processes, data stores, external actors, and trust boundaries.


\begin{table*}[!htbp]
\centering
\small
\caption{Impact scenarios for the OpenClaw deployment, derived from the business impact assessment (Appendix~\ref{sec:appendix-impact-matrix}).}
\begin{tabular}{c l l l l}
\toprule
\textbf{ID} & \textbf{Asset} & \textbf{CIA} & \textbf{Impact Scenario} & \textbf{Business Impact} \\
\midrule
IS1 & A1 Credentials & C & API keys exfiltrated & High \\
IS2 & A2 Messaging & I & Unauthorized messages sent to customers or contacts & High \\
IS3 & A3 Workspace & C & Source code / secrets leaked & High \\
IS4 & A4 Memory & I & Long-term memory poisoned & Moderate \\
IS5 & A6 Database & A & Booking platform and payments disrupted (heavy queries) & High \\
IS6 & A6 Database & C & Customer PII and booking records exfiltrated & High \\
IS7 & A5 Agentic App & I & Security controls weakened via config change & High \\
IS8 & A6 Database & I & Booking or invoice records corrupted & High \\
IS9 & A7 End Users & C/I & Customers phished via agent-sent messages & High \\
\bottomrule
\end{tabular}
\label{table:openclaw-impact-scenarios}
\end{table*}

\subsection{Business Impact Assessment Matrix}
\label{sec:appendix-impact-matrix}

Table~\ref{table:openclaw-impact-matrix} presents the stakeholder-assessed business impact for each asset and CIA dimension in the OpenClaw deployment. For availability, the impact is independent of the threat source, so a single rating is assigned. Cells marked ``---'' indicate combinations that are not applicable to this deployment. The matrix assesses impact only; the specific threat source's capability is introduced during attack tree construction (Section~\ref{subsec:attack-tree-construction}), where it determines the capability fit for each vector. Not every threat source is assessed against every impact scenario: if a threat source lacks credible intent or a viable targeting path for a given scenario, that combination is excluded from analysis (see Section~\ref{subsec:impact-assessment}).

\begin{table*}[!htbp]
\centering
\small
\caption{Business impact assessment matrix for the OpenClaw deployment.
  The adversarial column covers all adversarial threat sources, as the business impact
  of a given compromise is independent of the attacker's identity.}
\begin{tabular}{l l c c c}
\toprule
\textbf{Asset} & \textbf{CIA} & \textbf{Adversarial} & \textbf{Accidental} & \textbf{All (Avail.)} \\
\midrule
A1 Credentials    & C & High     & Low      & --- \\
                   & I & ---      & ---      & --- \\
                   & A & ---      & ---      & --- \\
\midrule
A2 Messaging       & C & ---      & ---      & --- \\
                   & I & High     & Moderate & --- \\
                   & A & ---      & ---      & --- \\
\midrule
A3 Workspace       & C & High     & ---      & --- \\
                   & I & High     & Moderate & --- \\
                   & A & ---      & ---      & --- \\
\midrule
A4 Memory          & C & ---      & ---      & --- \\
                   & I & Moderate & Moderate & --- \\
                   & A & ---      & ---      & --- \\
\midrule
A5 Agentic App     & C & ---      & ---      & --- \\
                   & I & High     & Moderate & --- \\
                   & A & ---      & ---      & Moderate \\
\midrule
A6 Database        & C & High     & ---      & --- \\
                   & I & High     & Moderate & --- \\
                   & A & ---      & ---      & High \\
\midrule
A7 End Users       & C & High     & ---      & --- \\
                   & I & High     & ---      & --- \\
                   & A & ---      & ---      & --- \\
\bottomrule
\end{tabular}
\label{table:openclaw-impact-matrix}
\end{table*}

\subsection{Attack Tree Construction Details}
\label{sec:appendix-attack-trees}

This appendix presents the full attack tree decomposition for the OpenClaw deployment, constructed in three stages following the process in Section~\ref{subsec:attack-tree-construction}. The decomposition proceeds from impacts to objectives (Figures~\ref{fig:openclaw-adv-impacts} and~\ref{fig:openclaw-nonadv-impacts}), objectives to techniques (Figure~\ref{fig:openclaw-obj-techniques}), and techniques to architecture-specific vectors (Figures~\ref{fig:openclaw-dpi-vectors}--\ref{fig:openclaw-hallucination-vectors}).

\subsubsection{From Impacts to Objectives}

\begin{figure*}[]
\centering
\begin{forest}
for tree={
  grow'=0,
  child anchor=west,
  parent anchor=east,
  edge path={\noexpand\path[\forestoption{edge}]
    (!u.parent anchor) -- +(5pt,0pt) |- (.child anchor)\forestoption{edge label};},
  l sep=10pt,
  s sep=3pt,
},
where level=0{draw=none, edge=none, inner sep=0pt, s sep=14pt}{},
where level=1{minimum width=62mm, text width=62mm, align=left}{},
[,phantom
  [\textbf{Impact:} Confidentiality \textit{(requires both)}
    [\textbf{Objective:} LLM Behavioural Manipulation]
    [\textbf{Objective:} Data Exfiltration]
  ]
  [\textbf{Impact:} Integrity \textit{(requires both)}
    [\textbf{Objective:} LLM Behavioural Manipulation]
    [\textbf{Objective:} Unauthorized Action Execution]
  ]
  [\textbf{Impact:} Availability \textit{(requires both)}
    [\textbf{Objective:} LLM Behavioural Manipulation]
    [\textbf{Objective:} Excessive Resource Consumption]
  ]
]
\end{forest}
\caption{Adversarial: Impact $\rightarrow$ Objectives (both required).}
\label{fig:openclaw-adv-impacts}
\end{figure*}
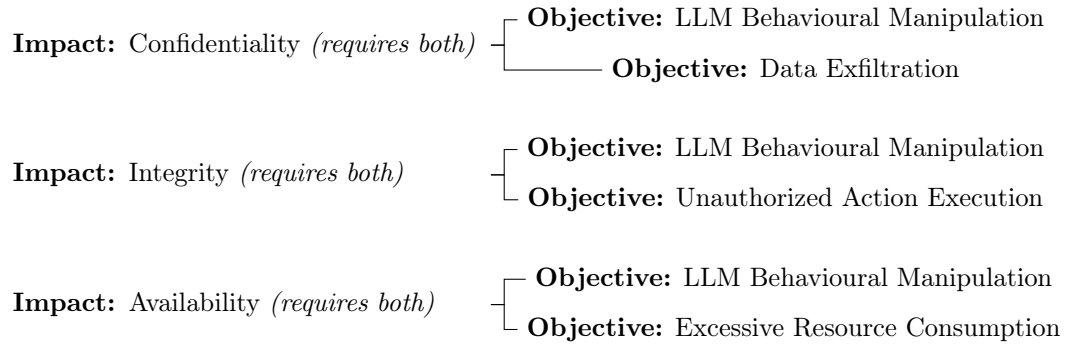

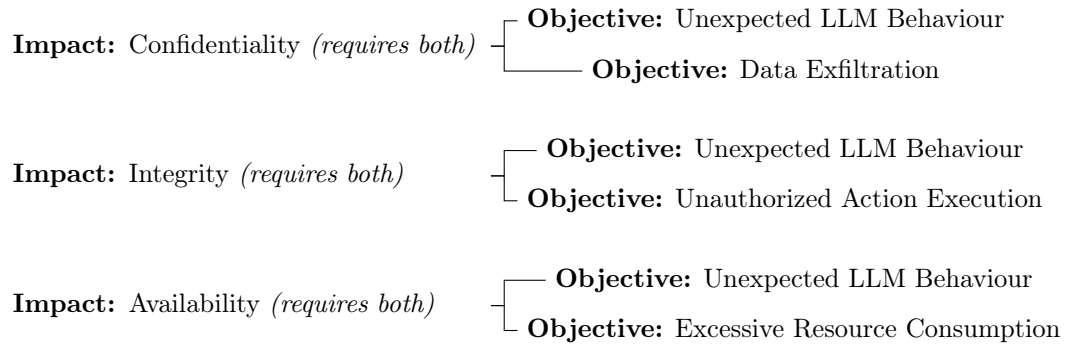
\begin{figure*}[]
\centering
\begin{forest}
for tree={
  grow'=0,
  child anchor=west,
  parent anchor=east,
  edge path={\noexpand\path[\forestoption{edge}]
    (!u.parent anchor) -- +(5pt,0pt) |- (.child anchor)\forestoption{edge label};},
  l sep=10pt,
  s sep=3pt,
},
where level=0{draw=none, edge=none, inner sep=0pt, s sep=14pt}{},
where level=1{minimum width=62mm, text width=62mm, align=left}{},
[,phantom
  [\textbf{Impact:} Confidentiality \textit{(requires both)}
    [\textbf{Objective:} Unexpected LLM Behaviour]
    [\textbf{Objective:} Data Exfiltration]
  ]
  [\textbf{Impact:} Integrity \textit{(requires both)}
    [\textbf{Objective:} Unexpected LLM Behaviour]
    [\textbf{Objective:} Unauthorized Action Execution]
  ]
  [\textbf{Impact:} Availability \textit{(requires both)}
    [\textbf{Objective:} Unexpected LLM Behaviour]
    [\textbf{Objective:} Excessive Resource Consumption]
  ]
]
\end{forest}
\caption{Non-adversarial: Impact $\rightarrow$ Objectives (both required).
  The decomposition mirrors the adversarial case, with ``Unexpected LLM Behaviour''
  replacing ``LLM Behavioural Manipulation'' as the first objective.}
\label{fig:openclaw-nonadv-impacts}
\end{figure*}

Figures~\ref{fig:openclaw-adv-impacts} and~\ref{fig:openclaw-nonadv-impacts} show the decomposition of adversarial and non-adversarial impact scenarios into their constituent objectives.

The integrity objective reflects OpenClaw's broad write access.
In read-only architectures, integrity impacts are limited to manipulating
the text returned to the user.
OpenClaw's single agent has full write access, i.e., shell execution, file I/O,
cross-channel messaging, and persistent memory, so integrity impacts extend
to system state changes such as modifying files, sending messages as the user,
corrupting database records, or poisoning the agent's long-term memory.

In OpenClaw's single-agent architecture, the non-adversarial path deserves
particular attention.
Because the agent autonomously executes tool calls without human confirmation,
the bottleneck for non-adversarial impacts shifts from action execution
(which happens automatically) to the hallucination probability itself.
The two-objective decomposition correctly captures that both the hallucination
\emph{and} the harmful action must co-occur for an impact to materialise.

\subsubsection{From Objectives to Techniques}

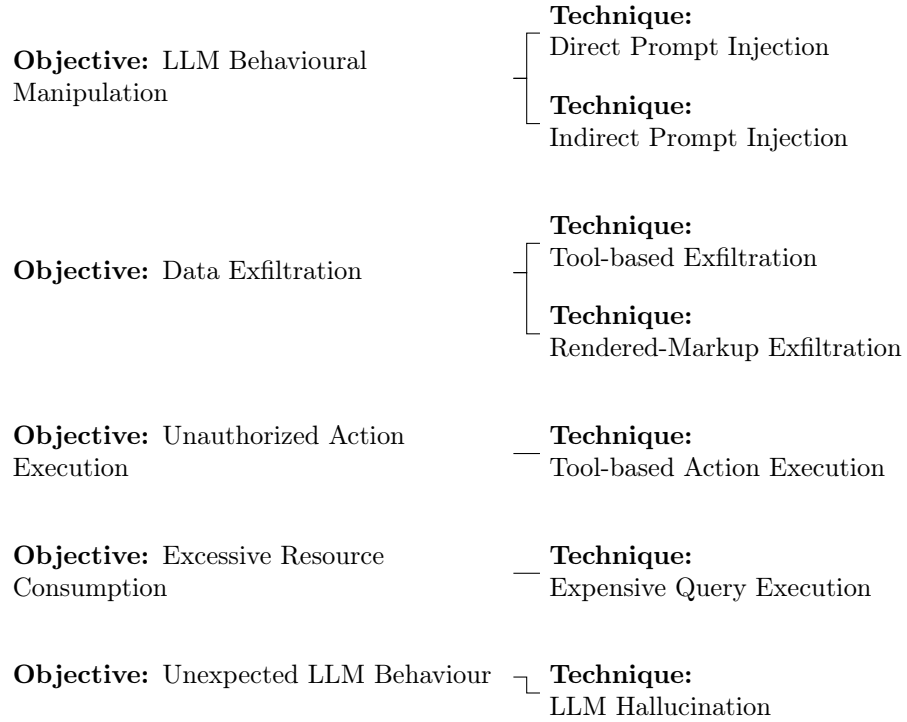
\begin{figure*}[]
\centering
\begin{forest}
for tree={
  grow'=0,
  child anchor=west,
  parent anchor=east,
  edge path={\noexpand\path[\forestoption{edge}]
    (!u.parent anchor) -- +(5pt,0pt) |- (.child anchor)\forestoption{edge label};},
  l sep=10pt,
  s sep=3pt,
},
where level=0{draw=none, edge=none, inner sep=0pt, s sep=14pt}{},
where level=1{minimum width=65mm, text width=65mm, align=left}{},
where level=2{
  text width=50mm,
  align=left,
  before typesetting nodes={
    content/.wrap pgfmath arg={
      \noexpand\textbf{Technique:}\\#1
    }{content()}
  }
}{}
[,phantom
  [\textbf{Objective:} LLM Behavioural \\ Manipulation
    [{Direct Prompt Injection}]
    [{Indirect Prompt Injection}]
  ]
  [\textbf{Objective:} Data Exfiltration
    [{Tool-based Exfiltration}]
    [{Rendered-Markup Exfiltration}]
  ]
  [\textbf{Objective:} Unauthorized Action \\ Execution
    [{Tool-based Action Execution}]
  ]
  [\textbf{Objective:} Excessive Resource \\ Consumption
    [{Expensive Query Execution}]
  ]
  [\textbf{Objective:} Unexpected LLM Behaviour
    [{LLM Hallucination}]
  ]
]
\end{forest}
\caption{Objectives $\rightarrow$ Techniques in the OpenClaw architecture.}
\label{fig:openclaw-obj-techniques}
\end{figure*}

Figure~\ref{fig:openclaw-obj-techniques} maps each objective to the attack techniques applicable in the OpenClaw architecture.

\subsubsection{From Techniques to Vectors}

\begin{figure*}[]
\centering
\begin{forest}
for tree={
  grow'=0,
  child anchor=west,
  parent anchor=east,
  edge path={\noexpand\path[\forestoption{edge}]
    (!u.parent anchor) -- +(5pt,0pt) |- (.child anchor)\forestoption{edge label};},
  l sep=10pt,
  s sep=3pt,
}
[\textbf{Technique:} Direct Prompt Injection
  [\textbf{Vector:} via Telegram direct message]
  [\textbf{Vector:} via email to agent]
]
\end{forest}
\caption{Attack vectors for \emph{Direct Prompt Injection} in OpenClaw.}
\label{fig:openclaw-dpi-vectors}
\end{figure*}
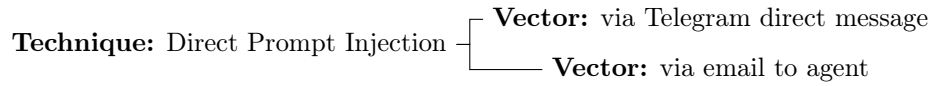

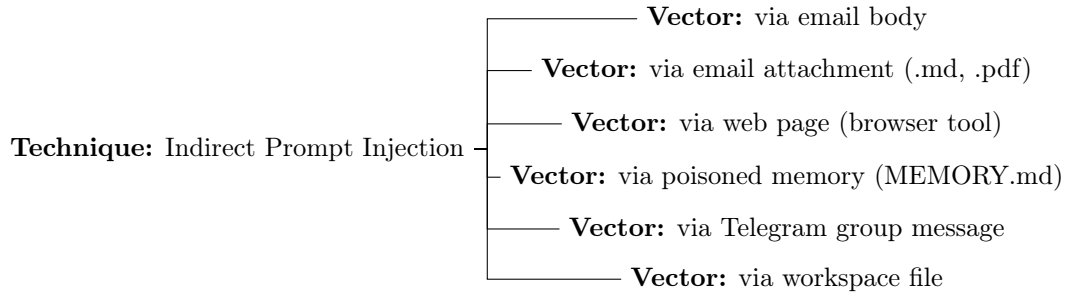
\begin{figure*}[]
\centering
\begin{forest}
for tree={
  grow'=0,
  child anchor=west,
  parent anchor=east,
  edge path={\noexpand\path[\forestoption{edge}]
    (!u.parent anchor) -- +(5pt,0pt) |- (.child anchor)\forestoption{edge label};},
  l sep=10pt,
  s sep=3pt,
}
[\textbf{Technique:} Indirect Prompt Injection
  [\textbf{Vector:} via email body]
  [{\textbf{Vector:} via email attachment (.md, .pdf)}]
  [\textbf{Vector:} via web page (browser tool)]
  [\textbf{Vector:} via poisoned memory (MEMORY.md)]
  [\textbf{Vector:} via Telegram group message]
  [\textbf{Vector:} via workspace file]
]
\end{forest}
\caption{Attack vectors for \emph{Indirect Prompt Injection} in OpenClaw.}
\label{fig:openclaw-ipi-vectors}
\end{figure*}

\begin{figure*}[]
\centering
\begin{forest}
for tree={
  grow'=0,
  child anchor=west,
  parent anchor=east,
  edge path={\noexpand\path[\forestoption{edge}]
    (!u.parent anchor) -- +(5pt,0pt) |- (.child anchor)\forestoption{edge label};},
  l sep=10pt,
  s sep=3pt,
}
[\textbf{Technique:} Expensive Query Execution
  [\textbf{Vector:} exec: \texttt{psql} with cartesian join]
  [\textbf{Vector:} exec: \texttt{psql} with full table scan]
  [\textbf{Vector:} exec: concurrent \texttt{psql} sessions]
]
\end{forest}
\caption{Attack vectors for \emph{Expensive Query Execution} targeting the PostgreSQL database.}
\label{fig:openclaw-query-vectors}
\end{figure*}
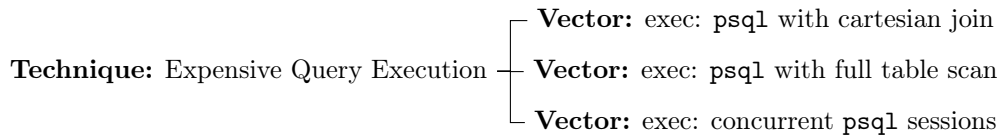

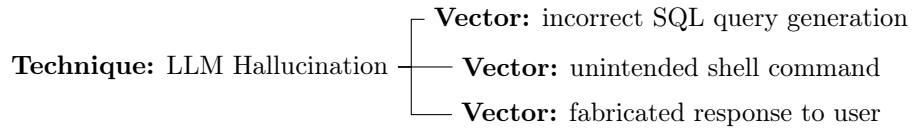
\begin{figure*}[]
\centering
\begin{forest}
for tree={
  grow'=0,
  child anchor=west,
  parent anchor=east,
  edge path={\noexpand\path[\forestoption{edge}]
    (!u.parent anchor) -- +(5pt,0pt) |- (.child anchor)\forestoption{edge label};},
  l sep=10pt,
  s sep=3pt,
}
[\textbf{Technique:} LLM Hallucination
  [\textbf{Vector:} incorrect SQL query generation]
  [\textbf{Vector:} unintended shell command]
  [\textbf{Vector:} fabricated response to user]
]
\end{forest}
\caption{Attack vectors for \emph{LLM Hallucination} in OpenClaw.}
\label{fig:openclaw-hallucination-vectors}
\end{figure*}

Figures~\ref{fig:openclaw-dpi-vectors}--\ref{fig:openclaw-hallucination-vectors} enumerate the concrete attack vectors for each technique, grounded in the OpenClaw architecture and its DFD (Figure~\ref{fig:openclaw-dfd-overview}).

\begin{figure*}[]
\centering
\begin{forest}
for tree={
  grow'=0,
  child anchor=west,
  parent anchor=east,
  edge path={\noexpand\path[\forestoption{edge}]
    (!u.parent anchor) -- +(5pt,0pt) |- (.child anchor)\forestoption{edge label};},
  l sep=10pt,
  s sep=3pt,
}
[\textbf{Technique:} Tool-based Exfiltration
  [\textbf{Vector:} exec: \texttt{curl} to attacker endpoint]
  [\textbf{Vector:} browser: HTTP request with encoded data]
  [\textbf{Vector:} message: email to attacker address]
]
\end{forest}
\caption{Attack vectors for \emph{Tool-based Exfiltration} in OpenClaw.}
\label{fig:openclaw-exfil-vectors}
\end{figure*}
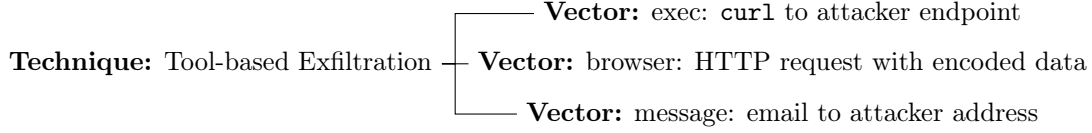

\begin{figure*}[]
\centering
\begin{forest}
for tree={
  grow'=0,
  child anchor=west,
  parent anchor=east,
  edge path={\noexpand\path[\forestoption{edge}]
    (!u.parent anchor) -- +(5pt,0pt) |- (.child anchor)\forestoption{edge label};},
  l sep=10pt,
  s sep=3pt,
}
[\textbf{Technique:} Rendered-Markup Exfiltration
  [\textbf{Vector:} Telegram: markdown image with data in URL (zero-click)]
  [\textbf{Vector:} Telegram: markdown link with data in URL (click required)]
]
\end{forest}
\caption{Attack vectors for \emph{Rendered-Markup Exfiltration} in OpenClaw.
  The agent encodes queried data in a URL embedded in its Telegram response.
  Telegram auto-fetches image URLs (zero-click); links require user interaction.}
\label{fig:openclaw-rendered-exfil-vectors}
\end{figure*}
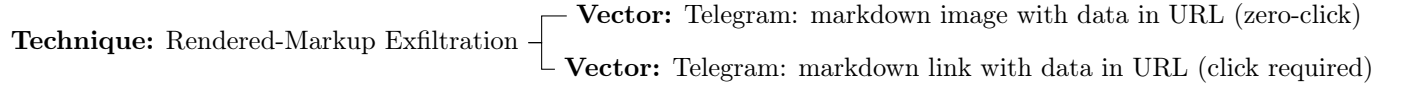

\begin{figure*}[]
\centering
\begin{forest}
for tree={
  grow'=0,
  child anchor=west,
  parent anchor=east,
  edge path={\noexpand\path[\forestoption{edge}]
    (!u.parent anchor) -- +(5pt,0pt) |- (.child anchor)\forestoption{edge label};},
  l sep=10pt,
  s sep=3pt,
}
[\textbf{Technique:} Tool-based Action Execution
  [\textbf{Vector:} exec: modify system configuration]
  [{\textbf{Vector:} exec: install persistence (crontab, startup scripts)}]
  [\textbf{Vector:} message: send phishing email/message as user]
  [\textbf{Vector:} file write: modify workspace or project files]
  [\textbf{Vector:} memory write: poison MEMORY.md or session logs]
]
\end{forest}
\caption{Attack vectors for \emph{Tool-based Action Execution} in OpenClaw.}
\label{fig:openclaw-action-vectors}
\end{figure*}
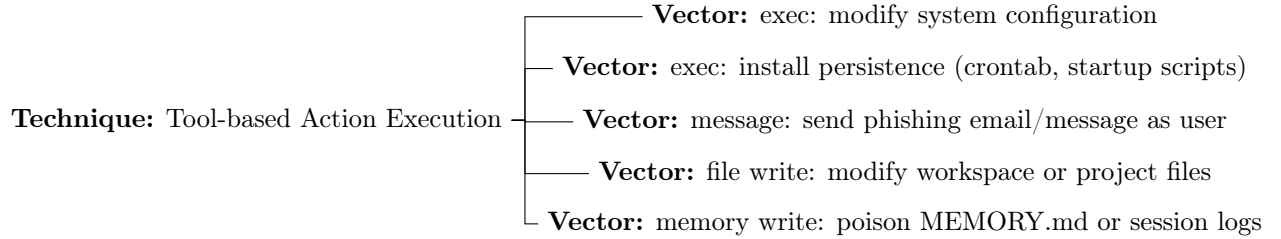

\subsection{Deep Dive: Database Risks---Availability and Integrity (IS5, IS8)}
\label{sec:appendix-deepdive-db}

\begin{figure*}[]
\centering
\resizebox{\linewidth}{!}{%
\begin{forest} for tree={align=left}
[\textbf{IS8: Booking records corrupted}\\
  \textcolor{purple}{\textbf{Default: Mod = min(Mod, High)}}\\
  \textcolor{blue!70!black}{\textbf{+ read-only DB role: Low = min(Mod, Low)}}
  [\textbf{Obj:} Unexpected LLM Behaviour\\
    \textcolor{purple}{\textbf{Mod = max(Mod)}}\\
    \textcolor{blue!70!black}{\textbf{Mod = max(Mod)}}
    [\textbf{Tech:} LLM Hallucination
      [\textbf{Vec:} incorrect SQL UPDATE/DELETE\\
        \textcolor{purple}{\textbf{Mod}} \footnotesize{(inherent: Mod $\to$ res: High $\to$ Mod)}\\
        \textcolor{blue!70!black}{\textbf{Mod}} \footnotesize{(inherent: Mod $\to$ res: High $\to$ Mod)}
      ]
    ]
  ]
  [\textbf{Obj:} Unauthorized Action Execution\\
    \textcolor{purple}{\textbf{High = max(High)}}\\
    \textcolor{blue!70!black}{\textbf{Low = max(Low)}}
    [\textbf{Tech:} Tool-based Action Execution
      [\textbf{Vec:} exec: \texttt{psql} with destructive query\\
        \textcolor{purple}{\textbf{High}} \footnotesize{(inherent: High $\to$ res: High $\to$ High)}\\
        \textcolor{blue!70!black}{\textbf{Low}} \footnotesize{(inherent: High $\to$ res: Low $\to$ Low)}
      ]
    ]
  ]
]
\end{forest}%
}
\caption{Attack tree for IS8 (database integrity, non-adversarial path),
  assessed for threat source: \emph{accidental} (non-adversarial).
  Vector labels follow the format \emph{inherent: manifestation likelihood $\to$ res: residual success likelihood $\to$ resulting likelihood}.
  For non-adversarial threats, inherent likelihood replaces capability fit; the inherent probability bounds the resulting likelihood regardless of residual success.
  Hallucination has an inherent Moderate probability; no control reduces it (res: High), but the result remains Moderate.
  In the default configuration, the agent has unrestricted database write access (res: High).
  A read-only database role blocks all write operations (res: Low), reducing the action execution objective from High to Low.}
\label{fig:openclaw-at-is8}
\end{figure*}
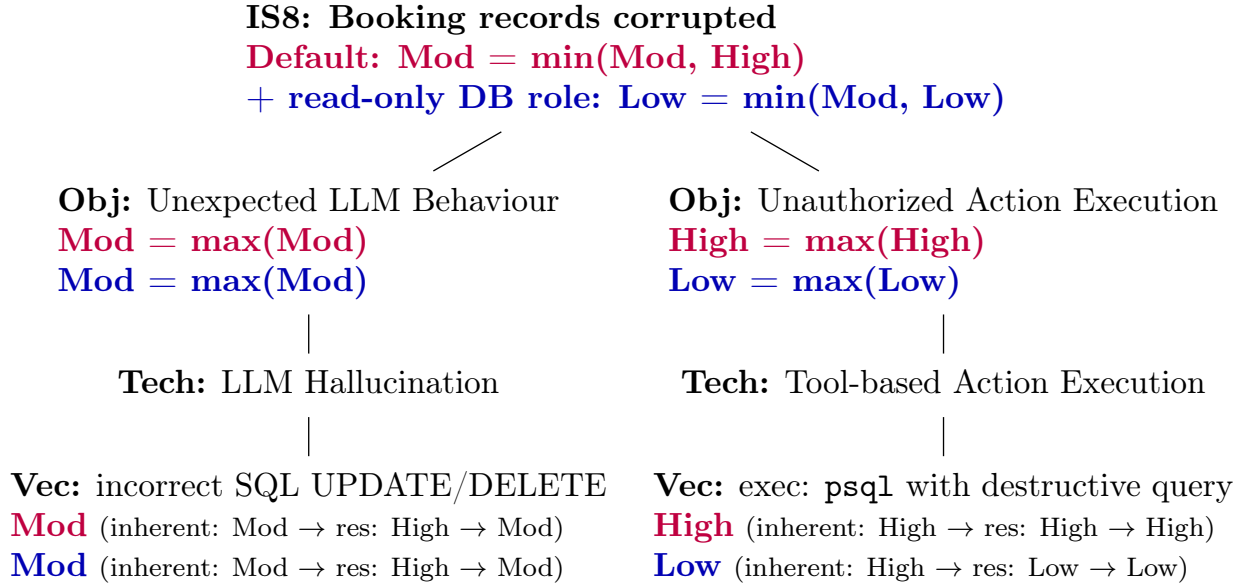

This deep dive examines two database-related impact scenarios that arise from
the agent's unrestricted SQL access via the \texttt{exec} tool (see Figures~\ref{fig:openclaw-at-is5} and~\ref{fig:openclaw-at-is8} for the full attack trees and Table~\ref{table:openclaw-is5-is8-risk} for the risk scores).
Because the production database is shared with the booking platform's web
application, payment processing worker, and reporting dashboard, any
disruption or corruption directly affects paying customers.

We present IS5 (availability) as a primarily \emph{adversarial} concern and
IS8 (integrity) as a realistic \emph{non-adversarial} risk.
This pairing illustrates how the same asset can be threatened through
fundamentally different paths, each requiring different controls.

\subsubsection{IS5: Booking Platform Disrupted (Adversarial)}

\paragraph{Objectives.}
Two objectives must succeed: \emph{LLM behavioural manipulation}
(to direct the agent to issue a specific query) and
\emph{excessive resource consumption} (the query must be expensive enough
to degrade availability for all connected services).

\paragraph{Narrative.}
An attacker places a crafted instruction in a web page that the developer
asks the agent to research.
The injected instruction directs the agent to execute a query such as
\texttt{SELECT * FROM bookings a CROSS JOIN bookings b}
against the production database via the \texttt{exec} tool (see Figure~\ref{fig:openclaw-query-vectors} for the full set of expensive query vectors).
This cartesian join produces an exponentially large result set, consuming
all available database connections and CPU, taking down the booking platform
for all customers.

\subsubsection{IS8: Booking Records Corrupted (Non-adversarial)}

\paragraph{Objectives.}
Two objectives must co-occur: \emph{unexpected LLM behaviour}
(the agent generating an incorrect write query) and
\emph{unauthorized action execution} (the query must actually modify data).
A hallucinated but syntactically invalid query, or one that is read-only,
does not cause an integrity impact; both conditions must be met.

\paragraph{Narrative.}
The developer asks the agent to ``mark all unpaid bookings older than 90~days
as cancelled.'' The agent generates a syntactically valid but semantically
incorrect query (see Figure~\ref{fig:openclaw-hallucination-vectors})---for example,
\texttt{UPDATE bookings SET status = 'cancelled' WHERE created\_at < now() - interval '90 days'}
omitting the \texttt{AND payment\_status = 'unpaid'} predicate.
The query cancels all bookings older than 90~days, including paid and
confirmed ones.
Customers receive erroneous cancellation notifications, the payment worker
attempts refunds for already-completed transactions, and the reporting
dashboard shows incorrect revenue figures.
No malicious input is required; the impact arises from the LLM's
imperfect understanding of the business logic encoded in the database schema.

\subsubsection{Full Attack Trees with Likelihood Scores}

\begin{figure*}[]
\centering
\resizebox{\linewidth}{!}{%
\begin{forest} for tree={align=left}
[\textbf{IS5: Booking platform disrupted}\\
  \textcolor{red}{\textbf{Default: High = min(High, High)}}\\
  \textcolor{blue!70!black}{\textbf{+ query guardrails: Low = min(High, Low)}}
  [\textbf{Obj:} LLM Behavioural Manipulation\\
    \textcolor{red}{\textbf{High = max(High, High)}}\\
    \textcolor{blue!70!black}{\textbf{High = max(High, High)}}
    [\textbf{Tech:} Indirect Prompt Injection
      [\textbf{Vec:} web page (browser)\\
        \textcolor{red}{\textbf{High}} \footnotesize{(capfit: High $\to$ res: High $\to$ High)}
      ]
      [\textbf{Vec:} email attachment\\
        \textcolor{red}{\textbf{High}} \footnotesize{(capfit: High $\to$ res: High $\to$ High)}
      ]
    ]
  ]
  [\textbf{Obj:} Excessive Resource Consumption\\
    \textcolor{red}{\textbf{High = max(High, High, High)}}\\
    \textcolor{blue!70!black}{\textbf{Low = max(Low, Low, Low)}}
    [\textbf{Tech:} Expensive Query Execution
      [\textbf{Vec:} cartesian join\\
        \textcolor{red}{\textbf{High}} \footnotesize{(capfit: Mod $\to$ res: High $\to$ High)}\\
        \textcolor{blue!70!black}{\textbf{Low}} \footnotesize{(capfit: Mod $\to$ res: Low $\to$ Low)}
      ]
      [\textbf{Vec:} full table scan\\
        \textcolor{red}{\textbf{High}} \footnotesize{(capfit: Mod $\to$ res: High $\to$ High)}\\
        \textcolor{blue!70!black}{\textbf{Low}} \footnotesize{(capfit: Mod $\to$ res: Low $\to$ Low)}
      ]
      [\textbf{Vec:} concurrent \texttt{psql} sessions\\
        \textcolor{red}{\textbf{High}} \footnotesize{(capfit: Mod $\to$ res: High $\to$ High)}\\
        \textcolor{blue!70!black}{\textbf{Low}} \footnotesize{(capfit: Mod $\to$ res: Low $\to$ Low)}
      ]
    ]
  ]
]
\end{forest}%
}
\caption{Attack tree for IS5 (database availability, adversarial path),
  assessed for threat source: \emph{competitor} (capability: Moderate).
  Vector labels follow the format \emph{capfit: capability fit $\to$ res: residual success likelihood $\to$ combined likelihood}
  (see Tables~\ref{table:capability-fit} and~\ref{table:combined-score}).
  Query guardrails reduce the resource consumption objective from High to Low:
  \texttt{statement\_timeout} and \texttt{EXPLAIN} cost validation cap query execution (res: Low for cartesian join and full table scan),
  while \texttt{max\_connections} limits prevent concurrent session exhaustion (res: Low for psql sessions).
  Injection vectors have no applicable controls (res: High).}
\label{fig:openclaw-at-is5}
\end{figure*}
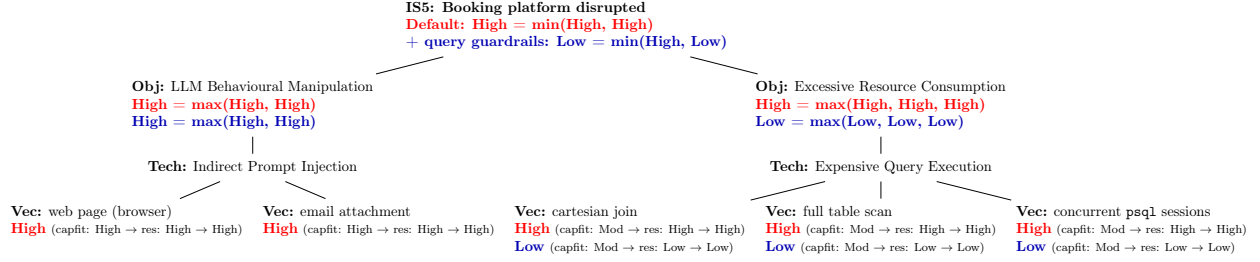

\subsubsection{Risk Assessment}

In the default configuration, IS5 reaches Very~High risk~(9): both the injection and resource consumption objectives are High, and the shared production database drives impact to High.
IS8 reaches High risk~(6) with Moderate likelihood, bottlenecked by the hallucination objective (Moderate inherent probability), while the action execution objective is High because the agent has unrestricted write access.

Despite arriving at different default risk levels, both scenarios benefit significantly from targeted controls, each reducing to Moderate~(3). The two scenarios require different controls because they are bottlenecked by different objectives.

\begin{table*}[]
\centering
\small
\caption{Risk scores for IS5 and IS8 under different control configurations.}
\begin{tabular}{l l l c c c}
\toprule
\textbf{Scenario} & \textbf{Threat Source} & \textbf{Configuration} & \textbf{Likelihood} & \textbf{Impact} & \textbf{Risk} \\
\midrule
IS5 & Competitor & Default & High & High & \cellcolor{red!50} Very High (9) \\
IS5 & Competitor & + query guardrails & Low & High & \cellcolor{yellow!50} Moderate (3) \\
\midrule
IS8 & Accidental & Default & Moderate & High & \cellcolor{orange!50} High (6) \\
IS8 & Accidental & + read-only DB role & Low & High & \cellcolor{yellow!50} Moderate (3) \\
\bottomrule
\end{tabular}
\label{table:openclaw-is5-is8-risk}
\end{table*}

\subsubsection{Control Effect Analysis}

IS5 and IS8 require \emph{different} database-level controls, illustrating
that the same asset can demand distinct mitigations depending on the CIA
dimension under threat.

\paragraph{IS5 (Availability).}
\begin{itemize}
  \item \textbf{Query timeout:} A PostgreSQL \texttt{statement\_timeout}
        caps resource consumption per query.
  \item \textbf{Query validation:} Running \texttt{EXPLAIN} on generated
        queries before execution and rejecting those above a cost threshold.
  \item \textbf{Connection limiting:} A PostgreSQL \texttt{max\_connections}
        setting or connection pool limit prevents concurrent session exhaustion.
\end{itemize}

\paragraph{IS8 (Integrity).}
\begin{itemize}
  \item \textbf{Read-only database user:} A read-only PostgreSQL role
        eliminates \texttt{UPDATE}, \texttt{DELETE}, and \texttt{INSERT},
        reducing the action execution objective from High to Low.
  \item \textbf{Transaction review:} Wrapping write queries in
        \texttt{BEGIN}/\texttt{ROLLBACK} with human confirmation before
        \texttt{COMMIT}.
\end{itemize}

Notably, \textbf{Docker sandboxing does not help} with either scenario; the
database must remain accessible for the tool to function.
IS6 (confidentiality) presents a parallel challenge: while sandboxing blocks
\texttt{exec}-based exfiltration, message-based and rendered-markup vectors
survive because they run outside the container.
All three scenarios illustrate that effective controls must operate at the
layer where the threat materialises, i.e., the database layer for IS5 and IS8,
and the messaging and output layer for IS6.

\end{document}